\documentclass[11pt,twocolumn]{article}
\usepackage[margin=1in]{geometry}
\usepackage{graphicx}
\usepackage{amsmath}
\usepackage{amssymb}
\usepackage{times}
\usepackage[numbers]{natbib}
\usepackage{booktabs}
\usepackage[hidelinks]{hyperref}
\usepackage{siunitx}

\title{Test-time Adaptation of Tiny Recursive Models}
\author{Ronan McGovern\\Trelis LTD\\\texttt{Trelis.com}}
\date{}

\begin{document}

\maketitle

\begin{abstract}
Prior to the close of the 2025 ARC Prize competition, the leading open source approach - known as TRM, or Tiny Recursive Models - involved training a 7M parameter recursive neural network on augmented variants of ARC tasks. That approach scored approximately 7.8\% on the public ARC AGI II evaluation set, but required a level of compute far in excess of what is allowed during the competition. This paper shows that, by starting from a tiny recursive model that has been pre-trained on public ARC tasks, one can efficiently fine-tune on competition tasks within the allowed compute limits. Specifically, a model was pre-trained on 1,280 public tasks for 700k+ optimizer steps over 48 hours on 4xH100 SXM GPUs to obtain a ~10\% score on the public evaluation set. That model was then post-trained in just 12,500 gradient steps during the competition to reach a score of 6.67\% on semi-private evaluation tasks. Notably, such post-training performance is achieved by full-fine tuning of the tiny model, not LoRA fine-tuning or fine-tuning of task embeddings alone.
\end{abstract}

\section{Introduction}

\subsection{Description and Brief History of the ARC Prize}
The ARC (Abstract Reasoning Challenge) Prize is a competition to efficiently solve unseen abstract tasks using computers. ARC tasks are designed to be solvable by humans but challenging for programmable approaches. The competition imposes a limit on the amount of compute that may be used to solve tasks, which is why approaches must not just generalise to unseen tasks but also solve them efficiently.

Historically, the leading approach to solving the first series of ARC tasks (ARC AGI I) was a brute force search - with heuristics - through a library of operations (e.g. flip, rotate, repattern) \cite{Chollet2024ARCPrize}. This was outperformed by a number of entries in the 2024 competition that involved pre-training a deep transformer model in the single billion parameter range. Such approaches involved pre-training on ARC training tasks prior to the competition, followed by post-training on test-time tasks within sandboxed competition environment with limited compute. These approaches lifted scores on ARC AGI I from roughly 20-30\% up into the 40\%+ range.

In 2025, ARC AGI II tasks were released of significantly higher difficulty than those in ARC AGI I. At the time of release, previous year's competition approaches scored in the low single digits, and leading large language models - including those far exceeding the compute limits of the competition - struggled to score beyond a few percent on the updated benchmark. Although, by late 2025, reasoning language models were scoring in the low double digits. 

\subsection{Tiny Recursive Models as a Basis for Solving ARC AGI II Tasks}

Compared to the ~1B to 7B parameter deep transformer models of 2024, which employed pre-training followed by competition-time fine-tuning, a new recursive model approach emerged in 2025 with the HRM paper \cite{Wang2025HierarchicalReasoning}, taking a number of new directions:
\begin{enumerate}
    \item A smaller model size, well less than 100M parameters
    \item The use of recursive layers rather than a single once-through deep network
    \item A focus on training from scratch on competition tasks, rather than first pre-training on ARC-like tasks
\end{enumerate}

This approach scored ~5\% on ARC AGI II tasks, which, while low in absolute terms, constituted the leading open source approach at the time, and was surprising for a model of such small size.

This HRM (Hierarchical Reasoning Model \cite{Wang2025HierarchicalReasoning}, hierarchical because it involves a hierarchy of recursive loops) was soon improved on in the Tiny Recursive Models (TRM) \cite{JolicoeurMartineau2025TinyRecursive} paper by reducing the architecture to a single neural net of 7M parameters, plus task embeddings. Although involving fewer model parameters, this updated approach improved in performance, reaching ~7.8\% on ARC AGI II tasks.

While at the time the leading open source approach, this TRM approach required pre-training on four H100 GPUs for over 48 hours - well in excess of the compute allowed in the ARC Prize 2025 competition. This gave motivation to the question of whether and how the tiny recursive model approach could be improved in compute efficiency, and made suitable for a competition entry.

\subsection{Fitting a Tiny Recursive Model Approach within Compute Requirements for the ARC Prize 2025 Competition}

The ARC Prize 2025 competition allows for the use of four L4 accelerators for twelve hours. TRM pre-training on 1,280 tasks for 100k epochs takes roughly 48 hours on 4xH100 SXM GPUs. Accounting for a factor of roughly 8x between bf16 flops on a H100 SXM versus an L4 accelerator, there is only about 1/8 x 1/4 = 1/32th of the compute necessary to complete full pre-training on this model size. Cast in different terms, pretraining with the TRM paper's approach takes approximately 750k optimizer steps with a global batch size of 768, while the competition runtime allows for only about 15k optimizer steps at a batch size of 384, given that one must also allow time for inference/evaluation.

Rather than train a model from scratch, an obvious approach is to instead fine-tune a model that has been pre-trained on public ARC tasks. There is no limit on the compute that can be used to pre-train such a model in advance of submission, although, the amount of publicly available ARC tasks, in particular tasks calibrated to ARC AGI II difficulty is very small. The question is, given competition tasks are new and unseen, and, given the model's capacity is small relative to pre-training and competition fine-tuning approaches of 2024, whether using a pre-trained TRM model can give a head start to fine-tuning at all?

\subsection{Approach and Contribution}
This paper demonstrates that starting from a pre-trained tiny recursive model significantly accelerates training on new, unseen tasks. The performance of such a fine-tuned model trends towards that of pre-training from scratch, although has not been found to quite reach or exceed the level of full pre-training performance - at least given the competition compute limits. Interestingly, the best performance was achieved by fully fine-tuning the pre-trained model, not by LoRA fine-tuning or by training the task id embeddings. Specifically, a pre-trained model achieving ~10\% on the ARC AGI II public evaluation split was capable of achieving 6.67\% when fine-tuned on unseen semi-private tasks in the Kaggle competition environment.

Attempts were also made to improve the performance by using a stronger pre-trained model - trained on more and/or higher quality data for longer. These attempts were unsuccessful, although it remains possible - or even likely - that different choices of architecture, dataset or training hyperparameters could still lead to better results.

\section{Methods}\label{sec:methods}

A recursive transformer model was pre-trained on ARC AGI II training tasks in close accordance to the Tiny Recursive Model paper \cite{JolicoeurMartineau2025TinyRecursive}. During competition submissions, this pre-trained model was fully fine-tuned on the train example pairs of the test tasks. This fine-tuned model was then used to predict test example outputs, using a majority voting method.

\subsection{Pre-training}
Three models were pre-trained. A first model was trained almost exactly in line with the original TRM paper. A second model was pre-trained with an expanded pre-training dataset and for double the original number of epochs. A third model was then pre-trained with a smaller dataset, filtered for tasks matching ARC AGI II public evaluation split difficulty.

\subsubsection{Original Paper Replication - 100k epochs}

A TRM was trained in close alignment with the original paper, but with only 4 lower reasoning cycles instead of 6 reported in the paper. This deviation was unintentional and resulted from a commit in the TRM GitHub repository using that same value. Interestingly, other ablations by Xin Gao \cite{Gao2025ARCAGI1TRM} and Konstantin Schürholt \cite{Schuerholt2025HRMAnalysis} also use this value of 4.

The data mix matched that of the original paper and included:
\begin{itemize}
    \item ARC AGI II Training split [1000 tasks]: train + test example pairs
    \item Concept ARC split [160 tasks]: train + test example pairs
    \item ARC AGI II Evaluation split [120 tasks]: train example pairs only (test used for evaluation)
\end{itemize}

\subsubsection{Extended Data for 200k Epochs}

Following the heuristic that neural nets often improve in performance with longer pre-training, a second model was pre-trained for double the original number of epochs.

Following the heuristic of more, higher quality, in-distribution data helping the performance of neural nets, the dataset was expanded in two ways:
\begin{enumerate}
    \item Inclusion of evaluation split test example pairs: Concretely, train example pairs from 110 tasks were included in pre-training and test example pairs from 100 tasks were included in pre-training. Test example pairs from 10 tasks were withheld for evaluation during pre-training, while train AND test example pairs from 10 tasks were withheld entirely for use in post-training.
    \item Inclusion of 50 tasks from Simon Strandgaard's "tama" dataset \cite{StrandgaardTamaDataset}. These human-reviewed tasks cover concepts typical in ARC challenges and are of a difficulty somewhere between ARC AGI I and ARC AGI II. The hope for including this data was to broaden and enhance the pre-training dataset.
\end{enumerate}

In sum, this meant pre-training on:
\begin{itemize}
    \item ARC AGI II Training split [1000 tasks]: train + test example pairs
    \item Concept ARC split [160 tasks]: train + test example pairs
    \item ARC AGI II Evaluation split [110 tasks]: train example pairs from all 110 tasks and test example pairs from 100 tasks (the other ten serve as evaluation during pre-training)
    \item tama [50 tasks]: train + test example pairs
\end{itemize}

While there is the advantage of a higher quality, more in-distribution and larger pre-training dataset, there is a large trade-off in being able to assess performance with evaluation and test hold-out sets of only 10 tasks each, particularly when assessing performance with a granularity of one percent and where the anticipated score is in the region of 1-10 percent.

\subsubsection{Filtered Hard Data for 1M Epochs}

This third pre-trained variant aimed to test the heuristic that it can be better to train neural nets on a smaller amount of higher quality data for more epochs than on more mixed-quality data for fewer epochs. The same model and hyperparameters were used but training on 110 tasks from the ARC AGI II evaluation split and 120 hard tasks from the ARC AGI II training split. Training tasks were determined to be hard based on the ability of GPT-5-mini to write a python program that successfully solved all train and test example pairs. Specifically, any task for which GPT-5-mini could write a correct program, given approximately eight attempts, was filtered out. This left 137 remaining ARC AGI II training tasks, from which 120 were selected as a traininghard split. This filtering is somewhat arbitrary and skewed both because it filters based on a) python programming performance and b) LLM, not human, performance. The method was used because of prior unreported work attempting to solve ARC tasks by writing python programs. The choice of GPT-5-mini as a model was made because it was capable of solving only a few ARC AGI II evaluation tasks by writing python programs. As such, if a task can be solved through python program writing by GPT-5-mini this correlates with the task being too easy for ARC AGI II level tasks. The motivation for this "hard" dataset split was to have tasks more representative of the semi-private evaluation set on which performance is ultimately graded for the 2025 competition.

In sum, this meant pre-training on 230 tasks:
\begin{itemize}
    \item ARC AGI II Training split, filtered with GPT-5-mini program writing for "hard" tasks [120 tasks]: train + test example pairs
    \item ARC AGI II Evaluation split [110 tasks]: train example pairs from all 110 tasks and test example pairs from 100 tasks (the other ten serve as evaluation during pre-training)
\end{itemize}

All data splits are available on the Trelis TRM fork on Github \cite{TrelisResearchTRM}.

\subsection{Post-training}
Four approaches to post-training were conducted:
\begin{enumerate}
    \item Full-fine tuning.
    \item Fine-tuning of embeddings only.
    \item Full-fine tuning, but training only embeddings for the first quarter of optimizer steps.
    \item LoRA + embeddings fine-tuning.
\end{enumerate}

All post-training runs were conducted in the same manner and with the same hyperparameters as the pre-training runs, but with the following differences:
\begin{enumerate}
    \item At the start of each post-training run, the model was initialised from a pre-trained model.
    \item Owing to lower total VRAM available on 4xL4s compared to 4xH100 SXMs, a global batch size of 384 was used instead of 768 for pre-training. Accordingly, the learning rate for the model trunk and for the embeddings were doubled (to 2e-4 and 2e-2, respectively).
    \item Owing to the compute limitation at competition time, training was run for only 15,000 optimizer steps, rather than $\sim$750k in the original pre-training.
    \item Post-training was conducted only on train examples from test tasks (either competition tasks OR withheld tasks from the ARC AGI II public evaluation split). No additional data was mixed in. For competition runs, this means running on semi-private test tasks, which can only be done by making a formal submission to the ARC Prize 2025 competition. One submission is allowed daily.
    \item TRM uses a majority voting method to make test output predictions, defaulting to a vote over 1,000 augmented versions of each task. Since compute time is limited during the competition, only 256 or 512 augmentations are used at evaluation time -- although 1,000 are still used during pre- and post-training. The number of augmentations does not affect training time because epochs are defined as epochs over a given task and its variants (for a given task, a variant is sampled at random during each epoch).
\end{enumerate}

\subsection{Simulating and Measuring Post-training Performance}

Simulating post-training performance is difficult, as there are only 120 public evaluation tasks for ARC AGI II, and - if one is to post-train on those tasks - those same tasks cannot be included in one's pre-training dataset. Three workarounds were attempted, with varying degrees of success.

\subsubsection{Post-training an ARC AGI I pre-trained model}
To assess the effectiveness of post training one can first pre-train a model on ARC AGI I tasks, i.e. the train and test example pairs from the 400 task training split, and the test pairs from the 400 task evaluation split. Such a model was already available on HuggingFace from Xin Gao \cite{Gao2025ARCAGI1TRM} and so that model was used.

This pre-trained model - which has only six tasks that also appear in the ARC AGI II public dataset (see Appendix~\ref{sec:appendix_data_splits}) - can then be post-trained on the train example pairs of the ARC AGI II evaluation split, and then used to make predictions of the corresponding test example pairs. Note that, for post-training to be effective, it is not a necessary condition that this approach show positive results, because ARC AGI I tasks are significantly easier than ARC AGI II tasks. As such, a model pre-trained with ARC AGI I may be too weak to provide a head start to post-training on ARC AGI II level tasks. As it turns out, the ARC AGI I pre-trained model IS strong enough to meaningfully help with post-training for ARC AGI II grade tasks.

\subsubsection{Post-training an ARC AGI II pre-trained model, but with 10 public eval tasks withheld as a test set}
As an attempt at simulating post-training performance on an ARC AGI II pre-trained model, one can run pretraining on ARC AGI II data, but, withholding 10 evaluation tasks as a test set for post-training.

This was done, but the signal from only ten test tasks is weak, noisy and uninformative.

\subsubsection{Post-training an ARC AGI II pre-trained model via a formal ARC AGI 2025 Submission}
A positive result from pre-training an ARC AGI I model on ARC AGI II data motivates taking an ARC AGI II pre-trained model (pre-trained on data including ARC AGI II public eval training tasks) and submitting that for post-training on the semi-private tasks in the competition. The assumption is that an ARC AGI II pre-trained model (i.e. trained on ARC AGI II public eval data) is more in-distribution for the semi-private eval tasks than a model pre-trained only on what are much easier ARC AGI I tasks. Unfortunately, one cannot take an ARC AGI II pre-trained model and easily simulate post-training because the train example pairs from evaluation tasks are already within the pre-training data.

For the purpose of the competition, there were three pre-trained models submitted (the same three listed in the pre-training section):
\begin{enumerate}
    \item The ARC AGI II pre-trained model (for 100k epochs). Here, post-training could not be simulated as it had been pre-trained on ARC AGI II evaluation train examples.
    \item An ARC AGI II pre-trained model (for 200k epochs, and with expanded data) - from which 10 tasks were withheld during pretraining. Here, post-training was simulated, but was too noisy to be informative given only 10 withheld tasks.
    \item An ARC AGI II pre-trained model (for 1M epochs, and with a dataset filtered to include only 230 ARC AGI II difficulty tasks) - from which 10 tasks were withheld during pretraining. Again, post-training was simulated but too noisy to be informative.
\end{enumerate}

Since all three pre-trained models were submitted to the competition for pre-training, semi-private results are reported for each.

\subsection{Post-training Methods}
Four post-training methods are considered and evaluated on an ARC AGI I pre-trained model using ARC AGI II tasks. Based on those results, the chosen approach for competition submissions was to fine-tune embeddings for one quarter of post-training epochs and then fully fine-tuning for the remaining epochs.

\subsubsection{Full fine-tuning}
Full fine-tuning involved updating all model parameters, as in pre-training.

\subsubsection{Fine-tuning of embeddings only.}
In TRM, each task and any augmented variant of that task is assigned a task id and has its own trainable embedding. At the start of post-training, the model sees a new set of tasks and variants, and so a new set of task ids and set of embeddings are initialised. These embeddings start the post-training process untrained, while the rest of the model parameters (attention, MLPs and heads) start in a pre-trained state.

As such, at the start of post-training, the task embeddings - which might be thought of as a description of the transformation involved in a given task variant - are out of sync with the model's trunk. With this as motivation, it appears reasonable to train only the embeddings to see whether they might be adapted by post-training in a manner that allows the model trunk to "execute" a program described by a newly trained task id. Were the training of embeddings sufficient, this might suggest that the trunk has the required tools to solve ARC tasks and the embedding need only be trained to signal to the trunk what tools should be used for the task at hand. There is a close analogy here with the Searching Latent Programs Spaces approach of Bonnet and MacFarlane \cite{Macfarlane2024SearchingLatent}, and more is discussed on that below.

\subsubsection{Fine-tuning of embeddings followed by full fine-tuning}
As it turns out, post-training only of the embeddings results in a score of near zero on held out tasks. This motivates the idea of first training only the embeddings for some epochs (one quarter of total post-training epochs) followed by full fine-tuning for the remainder of epochs.

\subsubsection{LoRA fine-tuning of the trunk plus embeddings fine-tuning}
Rather than train all parameters in the trunk (everything but the embeddings) one might instead train a set of low rank adapters for linear layers in the trunk. This reduces the number of parameters that are trainable. The motivation is potentially to avoid overfitting by training all parameters. However, TRMs are already low in parameter count, and the use of LoRA may reduce the ability of the model to adapt to the newly presented tasks. This appears to be the case.

\section{Results}

\subsection{Pre-Training}

All results for the TRM Paper ARC AGI II Replication Run are available at \url{https://wandb.ai/trelis/Arc2concept-aug-1000-ACT-torch},  see \texttt{pretrain\_att\_arc2concept\_4}, and the associated pre-trained model checkpoint is released at \cite{Trelis2025TRMARCAGII}.

Results for the extended training runs are available at \url{https://wandb.ai/trelis/Arc2-pretrain-final-ACT-torch}, and the released checkpoints are \cite{Trelis2025TRMARCAGIIAll200k} and \cite{Trelis2025TRMARCAGIIHard1M}.

Select results are shown here from the replication rather than the extended runs, because the evaluation set of ten tasks in the extended runs is too small to provide statistically meaningful insights on performance.

\begin{figure}[t]
    \centering
    \includegraphics[width=\columnwidth]{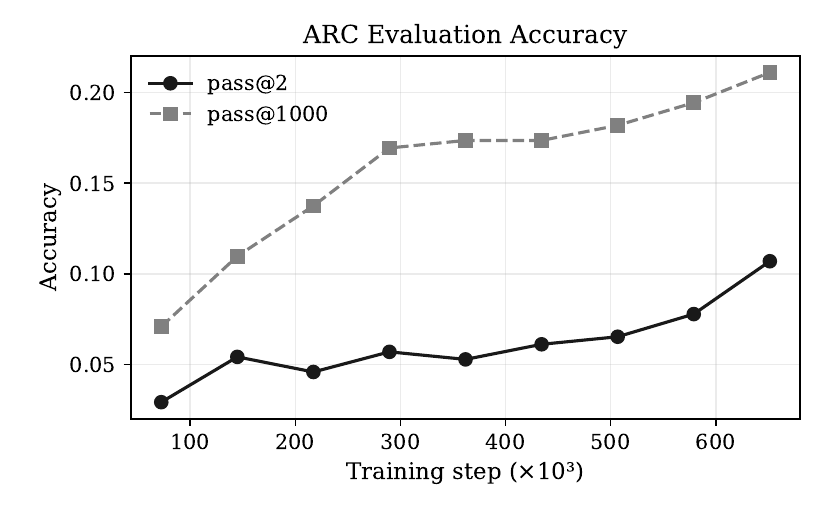}
    \caption{ARC evaluation pass@2 and pass@1000 accuracies throughout training for the \texttt{pretrain\_att\_arc2concept\_4} replication run.}
    \label{fig:pass_accuracy}
\end{figure}

\begin{figure}[t]
    \centering
    \includegraphics[width=\columnwidth]{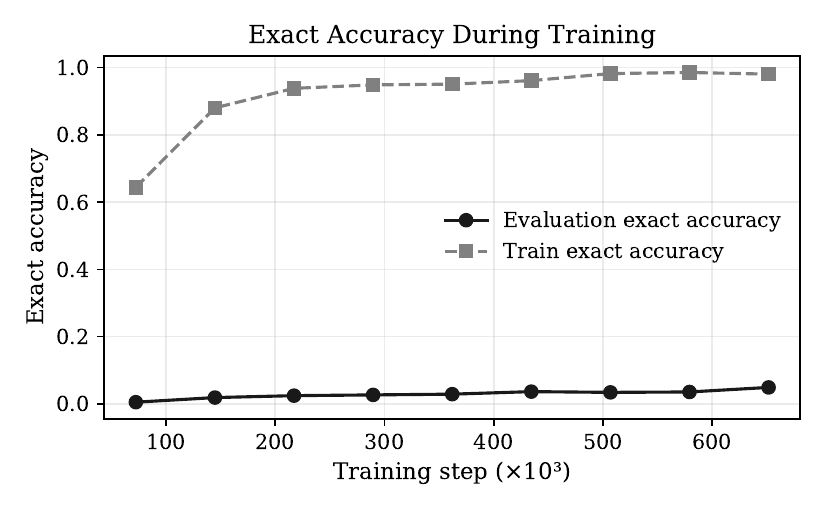}
    \caption{Evaluation and train exact accuracies throughout training for the \texttt{pretrain\_att\_arc2concept\_4} replication run.}
    \label{fig:exact_accuracy}
\end{figure}

Figure \ref{fig:pass_accuracy} shows the evolution of pass@2 and pass@1000 accuracy on the 120 ARC AGI II public eval tasks during the training run, reaching a final pass@2 score of ~10\%, which is higher than the originally reported score of 7.8\% in the TRM paper, presumably associated with stochastic behaviour during training. Figure \ref{fig:exact_accuracy} shows the per-example-pair exact accuracy (i.e. percentage of training pairs the model gets correct within a training batch of 768 example pairs) over the course of training. It is clear that the model heavily overfits the data, although continued training gradually brings up the exact accuracy on the evaluation set, although still far below 100\% by the end of training.

\subsection{Post-Training}
\subsubsection{Post-training of the ARC AGI I Pre-trained Model}
\begin{figure}[t]
    \centering
    \includegraphics[width=\columnwidth]{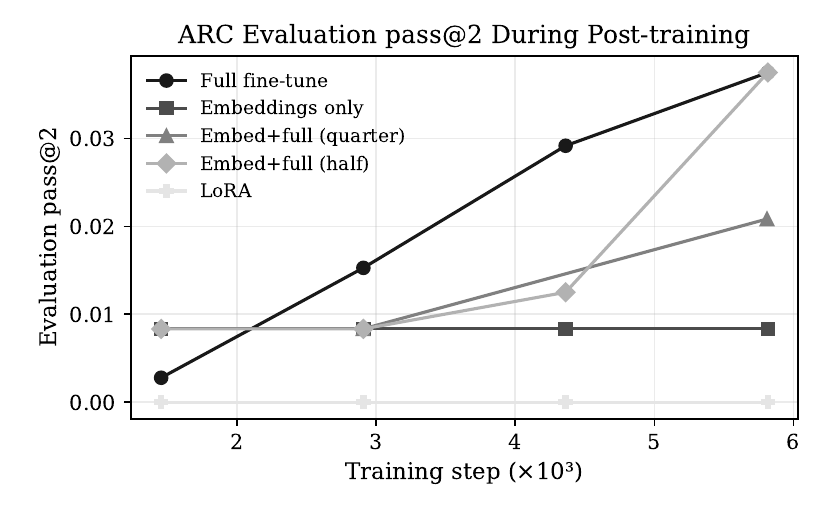}
    \caption{ARC evaluation pass@2 for the ARC AGI I post-training sweeps logged in \cite{Trelis2025PosttrainEval2}. Each curve shows the first $6$k optimisation steps for a different adaptation strategy.}
    \label{fig:posttrain_pass_accuracy}
\end{figure}

Post-training sweeps were conducted on the ARC AGI I checkpoint released by Xin Gao \cite{Gao2025ARCAGI1TRM}. All experiments were logged in the \texttt{Arc-eval2-aug-1000-ACT-torch} Weights \& Biases project \cite{Trelis2025PosttrainEval2}. Figure~\ref{fig:posttrain_pass_accuracy} compares evaluation pass@2 during the first $6$k optimisation steps for five adaptation strategies: full fine-tuning (\texttt{posttrain\_aa1\_aa2e}), embeddings-only (\texttt{posttrain\_aa1\_aa2e\_fe}), embeddings-only followed by full fine-tuning after a quarter (\texttt{posttrain\_aa1\_aa2e\_feq}) or half (\texttt{posttrain\_aa1\_aa2e\_feh}) of the steps, and LoRA+embeddings (\texttt{posttrain\_aa1\_aa2e\_lora}). Note that these evaluations are on the full ARC AGI II evaluation split, which unfortunately includes six contaminated tasks that also appear within the pre-training dataset - meaning that relative trends are perhaps of more significance than absolute pass@2 values.

Note that while only 6k optimizer steps were used for the above sweeps, competition submissions used either 12,000 or 15,000 optimizer steps.

\subsubsection{Post-training of ARC AGI II pre-trained models during competition submissions}
Post-training models pre-trained on ARC AGI II evaluation data provides limited insight because the ten evaluation tasks held out provided too small a sample to estimate performance reliably. The results are recorded at \cite{Trelis2025PosttrainEval2Holdout} and show 0\% pass@2 scores on the 10 task hold-out split after post-training on their train examples.

The three pre-trained models were submitted to the 2025 ARC Prize for post-training in the competition environment. All submissions reused the configuration in Section~\ref{sec:methods}, with the baseline TRM replication fine-tuned for $12.5$k steps and the extended variants for $15$k steps to stay within the twelve-hour budget. Table~\ref{tab:semi_private_scores} summarises the resulting semi-private evaluation scores. Re-submitting the same pre-trained model (the replication of the original TRM paper) yielded scores ranging from $3.33$ to $6.67$ depending on minor procedural tweaks (for example, adjusting the duration of frozen trunk updates or adding brief continued pre-training), underscoring the high stochastic variance inherent in pre-training, post-training and evaluation processes.

\begin{table}[t]
    \centering
    \caption{Semi-private evaluation accuracy achieved after post-training ARC AGI II checkpoints within the competition environment.}
    \label{tab:semi_private_scores}
    \begin{tabular}{@{}lS[table-format=1.2]@{}}
        \toprule
        Pre-trained model & {Accuracy (\%)} \\
        \midrule
        TRM paper replication & 6.67 \\
        Expanded data, 200k epochs & 4.25 \\
        Filtered hard data, 1M epochs & 1.27 \\
        \bottomrule
    \end{tabular}
\end{table}

\section{Discussion}

\subsection{Pre-training TRMs from scratch does not fit within competition compute limits}
Figure \ref{fig:pass_accuracy} shows the pass@2 and pass@1000 performance of a pre-trained - from scratch - tiny recursive model evaluated on the ARC AGI II public evaluation set of 120 tasks. For the purpose of the ARC Prize 2025 competition, it is the pass@2 score that counts. Entrants are allowed to submit two output grid predictions per test example, and the best output grid is the one that counts.

If one were to naively approach the pre-training-from-scratch approach of Fig.~\ref{fig:pass_accuracy} in the competition, there would only be sufficient compute to conduct about 10-20k optimizer steps. This corresponds to the very left of the plot, where the pass@2 score nears zero. Granted, the compute required scales with epochs over the data and the number of tasks in the dataset. While the TRM paper trains on ~1,280 tasks, one might consider pre-training - at competition time - on only the 240 competition tasks. This would reduce compute requirements by about 5x, although there are no guarantees of similar performance if the training data is reduced in diversity. Moreover, even a 5x reduction in the necessary optimizer steps per epoch with a smaller number of tasks - assuming one holds epochs constant - would result in needing 150k+ optimizer steps, still well beyond the 10-20k possible in the limited compute environment.

As such, the plot of pre-training performance in Fig.~\ref{fig:pass_accuracy} tells us that pre-training from scratch requires too much compute for the competition environment, if one is to follow the recipe of the TRM paper. Of course, one could try to find a different model, dataset or pre-training recipe that performs as well and works with much less compute. Some effort was spent on this, and is described in Section~\ref{sec:ablations}, but it was generally hard to match or exceed performance of the original TRM design, not to mention doing so with less compute.

\subsection{Fine-tuning a Pre-trained Model Significantly Accelerates Test-time Adaptation}
Figure \ref{fig:posttrain_pass_accuracy} is the fine-tuning analog of Fig.~\ref{fig:pass_accuracy} for pre-training. Notice how the number of optimizer steps to achieve an improvement in performance is orders of magnitude lower when starting from a pre-trained model than when pre-training from scratch.

The highest gains in accuracy are achieved by full fine-tuning OR by training only embeddings for a first portion of epochs (either half or one quarter), followed by full fine-tuning for the rest. Fine-tuning embeddings only OR LoRA fine-tuning (which additionally includes fine-tuning of embeddings) lead to low accuracy. There is a significant level of noise involved in these results. For example, LoRA+embeddings fine-tuning looks inferior to fine-tuning embeddings only in the pass@2 metrics of Fig.~\ref{fig:posttrain_pass_accuracy}. However, looking at a broader set of results - available in Weights and Biases \cite{Trelis2025PosttrainEval2} - reveals that LoRA+embeddings slightly outperforms embeddings-only when looking at pass@1000, a somewhat less noisy indicator of performance.

Clearly - and even with independent ARC tasks - it is possible to pre-shape the neural network in a manner that accelerates tuning on unseen tasks. This raises the question of what types of tasks are best used in pretraining to optimally shape the network for adaptation at competition time.

\subsection{It remains unclear what diversity of pre-training tasks best shapes a network for post-training on unseen tasks}
One heuristic in machine learning is that it can be better to train for longer on higher quality data than to expend that same amount of compute training for fewer epochs on a broader dataset of mixed quality.

Given ARC AGI II tasks are significantly harder than ARC AGI I tasks, and given ARC AGI II's public evaluation set is closely calibrated in difficulty to the semi-private and private evaluation sets, it was worth trying to pre-train only on ARC AGI II difficulty tasks, rather than the larger dataset of 1,000 ARC AGI II training tasks, most of which are far easier.

In practise, training for longer on this smaller but harder dataset (Filtered hard data, 1M epochs in Tab.~\ref{tab:semi_private_scores}) resulted in worse performance than pretraining on the larger original training dataset (TRM paper replication in Tab.~\ref{tab:semi_private_scores}). It is hard to know exactly why this is the case, and noise cannot be ruled out as a cause, but it suggests the ARC AGI II training set and/or the Concept ARC split of 160 tasks bring useful diversity or relevant concepts to the pre-training.

The other pre-training approach shown in Tab.~\ref{tab:semi_private_scores}, "Expanded data, 200k" involved training for longer than the original TRM paper, including test examples from ARC AGI II public evaluation tasks, and including additional concepts from the tama dataset \cite{StrandgaardTamaDataset}, all in the hope that added data diversity and training time might help with performance. Here, the difference in performance with the original TRM baseline (4.25\% vs 6.67\%) is likely within noise. Perhaps it is not surprising there is little improvement due to the expanded dataset because the dataset is only 50 tasks larger than the original TRM paper. Tama dataset concepts may overlap with concepts already present in ARC AGI II training data, and the inclusion of test examples from ARC AGI II evaluation data does not add new tasks, but only adds more examples per task.

\subsection{On the continued pre-training of pre-trained models}
The trend of pass@2 and pass@1000 scores in Fig.~\ref{fig:pass_accuracy} suggests that simply training for longer could lead to improved results. There is an important nuance with TRM's design, whereby task embeddings are trained for an index of tasks. If one does not save that mapping of tasks to embeddings, then one has no choice but to reinitialise and re-train the embeddings. Unfortunately, for the replication run (although not for the expanded data or filtered hard data pretraining runs) the embedding to task mappings were not saved (which involves saving the task and task augmentation dataset), and strict continued pre-training was not possible.

Nonetheless, continued pre-training was tried for a further ~72k optimizer steps (roughly 10\% more steps) - with reinitialisation of embeddings - but this led to lower submission performance of 3.33\% compared to 6.67\%. As such, reinitialisation of embeddings may lead to partial model collapse and/or overfitting, perhaps as the embeddings learn more grid specifics rather than the general transformations.

\subsection{On the use of augmentation-specific task embeddings}

Both HRM and TRM assign a unique embedding not just to each task, but to every augmented variant (flips, rotations, re-colours) of each task. From the model's perspective, each task variant is an entirely separate task. The model may or may not learn that these tasks are related.

It is interesting to ask whether:
\begin{enumerate}
    \item the model does learn that variants of the same task are related,
    \item it would be more efficient to use the same task\_id/embedding for all variants of a given task OR whether treating variants as independent tasks provides the model with some useful form of regularisation/generality during pre-training.
\end{enumerate}

\subsubsection{Using cosine similarity to measure the model's ability to encode related tasks}

As a measure of task relationship learning, one might look at the cosine of the angle between task embeddings for i) variants of the same task and ii) between the base/original/unvaried form of different tasks. If the TRM learns to encode task variants similarly, perhaps one should expect a rise in the cosine between embeddings during pre- and post-training. With this as motivation, these measures were recorded during the pretraining of the TRM for 200k epochs on the extended dataset (extended to include 'tama' and ARC EVAL II evaluation test examples).

\begin{figure}[t]
    \centering
    \includegraphics[width=\columnwidth]{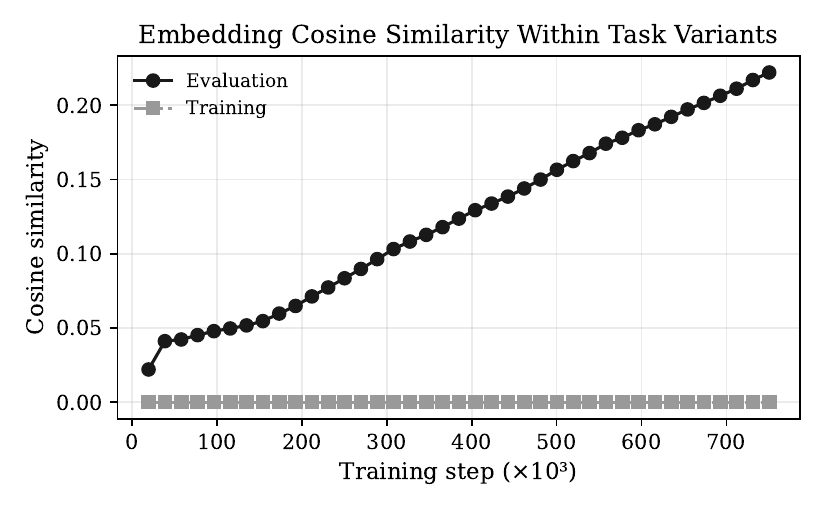}
    \caption{Embedding cosine similarity among augmented variants of the same task during extended pre-training.}
    \label{fig:embedding_cosine_within}
\end{figure}

Figure \ref{fig:embedding_cosine_within} shows the evolution of the cosine of the angle between task embeddings of variants of the same task (averaged across tasks in a given batch), for training and evaluation sets. Note that the angle between task id embeddings is reported as zero at most steps because there is at most one variant of each task per batch of 768 example pairs. If one looks in greater granularity at the data in Weights and Biases \cite{Trelis2025PretrainFinalRuns}, there are timesteps reaching cosine similarity of up to about 0.1 . This is more apparent in the training of the third model on the subset of hard tasks (not shown in these plots), since there are only 230 base task ids and every batch therefore includes more than one variant for each task. Nonetheless, the cosine similarity of task id embeddings for variants of the same task is low when measured on the training set. The cosine similarity is higher, and rising throughout training, when measured on the evaluation set. It is not obvious why training and evaluation sets diverge here and why the model might adapt to develop more similarity on evaluation example pairs. 

\begin{figure}[t]
    \centering
    \includegraphics[width=\columnwidth]{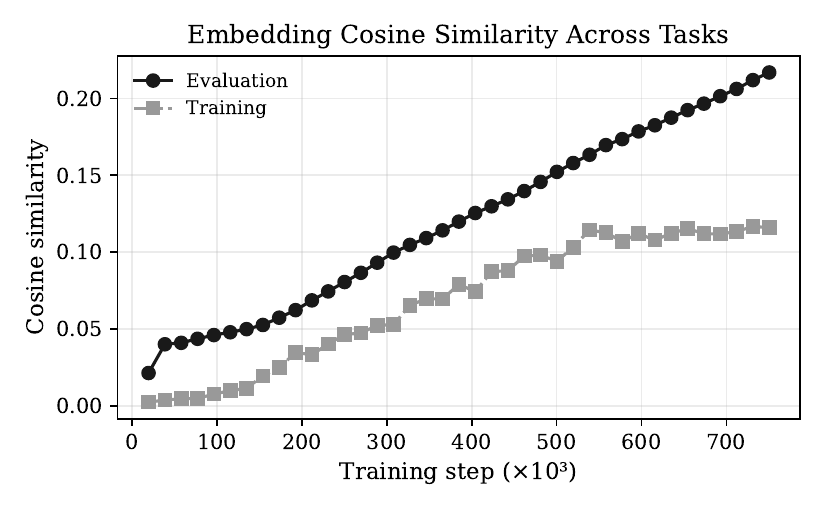}
    \caption{Embedding cosine similarity across base tasks for the extended pre-training runs.}
    \label{fig:embedding_cosine_across}
\end{figure}

Figure \ref{fig:embedding_cosine_across} illustrates the evolution - over the course of training - of the cosine of the angle between the base variants of different tasks within the same training batch. Although there is some gap between the cosine on training and evaluation examples, both rise together, although the measurement seems to asymptote more on the training examples. Interestingly, comparing Fig.~\ref{fig:embedding_cosine_across} and \ref{fig:embedding_cosine_within}, task id embeddings appear similarly distant within tasks as between tasks, suggesting the relationship between variants of the same task are not clearly expressed - at least within the embeddings alone.

\subsubsection{Pre-training a model with explicitly encoded embeddings}
Rather than assign an independent task id embedding to each augmented version of a task, one might instead assign the same task id to all variants of a given task, but then encode the augmentation type.

For example, one might simply encode flips and rotations by enumerating and embedding the 8 dihedral (d4 group) variants. To encode re-colours, one might build a look-up table covering the colour mapping pairs and train embeddings for that look-up table. In principle, this level of entropy can be captured in a single low-dimension embedding, perhaps no larger than 256, or even 128 dimensions.

This approach has the benefit of greatly reducing the number of model parameters by avoiding the need for a 512-dimension embedding for each task. While the TRM paper model has 7M parameters in its trunk, there are ~2.5 GB of parameters required to capture 1,000 augmentations of ~1,000 tasks (1k tasks x 1k augs/task x 512 dimensions = 512 M parameters). As such, when encoding each task augmentation individually, the TRM is more a 500M+ parameter model than a 7M parameter model. With explicit augmentation encodings, the size of the embeddings drops to 1k tasks x 512 dimensions = 512k parameters.

One might further expect that explicitly encoding augmentations makes it easier for the model to learn, and potentially require less compute to reach convergence. However, this proved not to be the case, and the performance of models trained with explicit encodings for embeddings (detailed in the slim-in and base-in branches of the Trelis fork of TRM \cite{TrelisResearchTRM}) was inferior to augmentation specific task embeddings (see Weights and Biases Project: https://wandb.ai/trelis/Arc2ethard-aug-1000-ACT-torch).

Perhaps the use of augmentation-specific task embeddings forces the model to generalise in a useful manner, but, then why isn't there stronger evidence of this in rising cosine similarity between in-task embeddings during pre-training?

\subsection{Post-training Tiny Recursive Models as a Variant of Searching Latent Program Space (SLPS)}

In their paper, Searching Latent Program Spaces, McFarlane and Bonnet \cite{Macfarlane2024SearchingLatent} describe an approach not dissimilar to that involved in post-training a tiny recursive model. They cast the problem of solving ARC tasks as a search for a vector embedding that describes the transformation involved in an ARC task, that program (or instruction) then being fed to a neural net and executed on a grid input.

All of the steps of post-training a TRM are there in Searching Latent Program Space:
\begin{enumerate}
    \item There is the pretraining of a neural net on ARC tasks, to establish a landscape for that category of problems.
    \item There is the post-training/fine-tuning on train examples from the test tasks, involving the search for the best latent (embedding) to describe the transformation at hand.
\end{enumerate}

Perhaps the performance gap between SLPS and TRM with post-training can be attributed to:
\begin{enumerate}
    \item The more complete set of augmentations used by TRM (dihedral group variants, recolours and translational augmentations).
    \item The fact that SLPS searches only for the best latent (i.e. trains only what might be thought of as the embeddings in TRM, but not other parameters). As such, SLPS finds the best combination of pre-trained primitives, but cannot add/encode new primitives.
    \item The recursive nature of the TRM's neural net, which allows for greatly increased effective depth while maintaining stability during training.
\end{enumerate}

\subsection{On the distribution of ARC AGI II tasks}

A major challenge in ARC AGI II is that there is little public data that is clearly in the distribution of the eventual ARC AGI II semi-private dataset. It is known that any approach scores remarkably similarly on the ARC AGI II public evaluation set and on the ARC AGI II semi-private (and likely private?) dataset. This means that those datasets are closely in-distribution. By contrast, the ARC AGI II training dataset (and the hard split) appears not to be in distribution as scores do not correlate closely with ARC AGI II evaluation sets.

For the purpose of research, it would have been highly useful to have an ARC AGI II training split of 120 tasks in the same distribution as the public eval and semi-private eval set. This would have allowed for pre-training on such a set, followed by post-training on the public eval set to accurately assess performance.

The closest approximation of this would perhaps have been to pre-train on 60 of the 120 public eval tasks, and post-train on the other 60. However, this has two drawbacks:
\begin{itemize}
    \item Statistical power, already small at just 120 tasks, is even smaller when one takes a subsplit.
    \item Model performance is sensitive to the amount of data that must be encoded. For the same model size, one cannot directly compare the performance pre or post training on 120 tasks versus 60 tasks.
\end{itemize}

\section{Other Ablations}\label{sec:ablations}
Other ablations are reported here, rather informally. Where available, links to Weights and Biases reports are provided.

\subsection{On the method of embedding initialisation in post-training}
When faced with unseen tasks, the TRM code base by default initialises new embeddings to the mean of pre-trained embeddings.

An ablation was run to instead initialise embeddings to a Gaussian norm with the mean and variance of the pre-trained embeddings. The hope was that by adding noise to the initialisation, the model may overfit less during post-training. However, this hurt performance.

\subsection{On the variation of batch size during post training}
Another heuristic in machine learning is that reducing batch size can sometimes introduce favourable noise and regularization that improves test performance.

While the TRM pre-training replication was run with the same batch size (768) as the original paper and on 4xH100 SXM, the other two pre-training runs were conducted at a global batch size of 1536 and with the learning rate doubled in order to make full use of 8xH100 compute cores, and cut training time in half. It is possible this had an adverse effect on pass@2 performance.

For post-training, ablations were run at smaller batch sizes of 96 and 32 (compared to 384) in the hope of improving regularisation. However, performance was not improved, and, post-training takes longer per epoch as CUDA core utilisation is degraded. A selection of post-training results are visible here: https://wandb.ai/trelis/Arc-eval2-aug-1000-ACT-torch .

\subsection{On improvements to majority voting}
The TRM code base uses a simple form of majority voting among augmentations for a given task in order to rank top predictions for a task example. TRM includes a halting head designed to indicate whether a task has been solved (which primarily serves to stop recursions early during training). However, for ARC AGI II tasks, although the halting head does learn to detected solved train examples, it does not learn to effectively detect solved evaluation test examples. For this reason, unlike in Sudoku tasks, the halting head cannot (yet) be effectively used to improve voting among predictions from augmentations.

\subsection{On the benefit of joint training on multiple tasks for compute efficiency}

If post training requires fewer optimizer steps than training from scratch -- to reach the same performance -- then clearly there is meaningful inter-task learning OR at least there are shared concepts involved in training TRM.

The effect can also be understood by pretraining on a single task versus 8 versus 120 tasks on a model of the same size. While it appears possible to achieve similar performance when training on a task individually, or combined with other tasks, it is more efficient -- again for a model of fixed size -- to jointly train on multiple tasks \cite{Trelis2025ArcEval2Clean}, showing that decomposing a post-training run into separate batches of post-training and inference does not materially improve performance but does increase total training time. Said differently, one can reach the same performance with less compute if one jointly trains on multiple tasks. As such, there appears to be joint concepts to be learned. It is possible of course that much of this joint learning involves primitive concepts relating to grid sizes and colours, as opposed to transformations themselves.

\section{On the use of smaller or larger model parameter counts}
In the hope of reducing pre-training requirements, ablations were run where the characteristic dimension of the model, and of task id embeddings, was reduced from 512 down to 256 - resulting in a model of roughly one quarter of the original parameter count. The code is available in the 'slim' branch of the Trelis TRM fork \cite{TrelisResearchTRM}, and results are shown in https://wandb.ai/trelis/Arc2ethard-aug-1000-ACT-torch . Smaller models were capable of scoring in the low single digits on ARC AGI II evaluation tasks, when compared on an iso-compute basis to the base model. However, there was not enough clear advantage to justify pursuing the direction further.

One ablation with a larger (1024 characteristic dimension) model was also run and is reported here: https://wandb.ai/trelis/Arc2concept-aug-1000-ACT-torch . It appeared to trend similarly in pass@2 evaluation score to the base model on an iso-compute basis, but a longer and more thorough run is required to make firm conclusions.

The limited ablations conducted leave unanswered the question of what the optimal model dimension should be, and what the optimal ratio of task embedding dimension relative to model trunk dimension (default ratio of 1) should be.

Intuitively, the size of the task embeddings relative to the model trunk dimension should be reflective of the relative entropy within task transformations versus that required to "execute" such tasks in the model trunk. Certainly the TRM design is heavily weighted towards storing information in the task embeddings because there is a 512 dimension embedding not just for every task but also for every augmentation. For 1k tasks one has 500M+ parameters for embeddings and just 7M for the trunk.

\section{Conclusion and future work}

Full fine-tuning allowed a pre-trained tiny recursive transformer model to be efficiently adapted in the compute-limited environment of the ARC AGI II competition. Effective adaptation appears to require updating both task id embeddings AND the trunk of the model at competition time.

Currently, the use of tiny recursive models has been limited to single or low double-digit scores on ARC AGI II tasks. While post-training of a pre-trained TRM is much more compute efficient than pre-training from scratch, it has not been shown that post-training can exceed or reach the performance achieved in pre-training. That pass@1000 metrics reach well above pass@2 metrics, often above 20\% on ARC AGI II difficulty tasks, suggests that performance improvements are possible with better pre-training. It remains an open question how far performance could be pushed through model size or hyper parameter improvements with or without further data additions or augmentations.

Future work may consider:
\begin{enumerate}
    \item Closer visual inspection of the tasks solved versus not solved by TRM-type approaches. Is it conceivable that tasks are solved owing to the types of augmentations (rotate, flip, translate, re-colour) that are applied? If so, are there other augmentations (e.g. shearing, re-patterning) that might assist in solving more tasks?
    \item What size model (specifically, size of hidden dimension) is most compute efficient for pretraining on a fixed number of tasks? Chinchilla laws for LLMs suggest that larger models may be more compute efficient (on a training compute basis) than smaller models. Such ablations were tried but, given the compute budget, it was not possible to reach a firm conclusion on the best model size. Note that - at the current TRM model size - inference/evaluation accounts for roughly 1 hour out of the 12 hours of compute time allowed. There is perhaps room for model size to be increased a little and evaluation prolonged at the expense of shorter post training. As it is, either 256 or 512 augmentations are used for inference, and there is little difference in performance between the two. As such, for the current model size competition-time training and inference are possibly not too far from optimal. The optimal split for larger, or smaller models, needs more analysis.
    \item For a constant size model hidden dimension, does increasing or decreasing the embedding dimension improve performance. Currently, it appears that the embedding dimension is not capturing similarities between task variants. Does this suggest the task id embedding is too large or too small?
    \item Improved optimisation of hyperparameters, especially the number of higher and lower loops that control effective model depth. Does increasing the level of recursion increase or decrease compute efficiency when holding the effective depth constant?
    \item Currently, at test time, the halting signal is ineffective, i.e. while the model learns to halt at early iterations on training data, it does not effectively halt on evaluation data. Are there ways to improve the halting behaviour OR are the tasks fundamentally too hard, perhaps suggesting that an even larger effective depth is required for training?
\end{enumerate}

\section{Acknowledgements}

This work was supported by Runpod, which provided \$15k of compute, and by Lambda Labs, which provided \$1k of compute. Thanks to Lewis Hemens for months of collaboration on ARC Prize research and support for compute costs. Thanks to Jack Boylan for assistance in running the pre-training replication.

\bibliographystyle{plainnat}
\bibliography{references}

\clearpage
\onecolumn
\appendix
\section{ARC Task Example Data Splits}\label{sec:appendix_data_splits}

\begin{table}[t]
    \centering
    \small
    \caption{Raw dataset splits used across experiments.}
    \label{tab:raw_splits}
    \begin{tabular}{lrrrr}
        \toprule
        \textbf{Challenges file} & \textbf{Puzzles} & \textbf{Avg. train inputs} & \textbf{Avg. test inputs} \\
        \midrule
        \texttt{arc-agi\_concept\_challenges.json} & 160 & 2.67 & 3.00 \\
        \texttt{arc-agi\_training2\_challenges.json} & 1000 & 3.23 & 1.08 \\
        \texttt{arc-agi\_evaluation2\_challenges.json} & 120 & 2.98 & 1.43 \\
        \texttt{arc-agi\_tama\_challenges.json} & 50 & 3.18 & 1.52 \\
        \texttt{arc-agi\_test\_challenges.json} & 240 & 3.20 & 1.08 \\
        \bottomrule
    \end{tabular}
\end{table}

\begin{table}[t]
    \centering
    \small
    \caption{Derived splits constructed for extended pre-training and evaluation.}
    \label{tab:derived_splits}
    \begin{tabular}{lrrrr}
        \toprule
        \textbf{Challenges file} & \textbf{Puzzles} & \textbf{Avg. train inputs} & \textbf{Avg. test inputs} \\
        \midrule
        \texttt{arc-agi\_evaluation2train\_challenges.json} & 100 & 2.96 & 1.44 \\
        \texttt{arc-agi\_evaluation2eval\_challenges.json} & 10 & 2.90 & 1.60 \\
        \texttt{arc-agi\_evaluation2test\_challenges.json} & 10 & 3.30 & 1.20 \\
        \texttt{arc-agi\_traininghard\_challenges.json} & 120 & 2.98 & 1.09 \\
        \texttt{arc-agi\_evaluation2clean\_challenges.json} & 114 & 2.97 & 1.46 \\
        \bottomrule
    \end{tabular}
\end{table}

\paragraph{Notes on derived splits.} The \texttt{evaluation2train}, \texttt{evaluation2eval}, and \texttt{evaluation2test} files are all sampled from the ARC AGI II evaluation split. These subsets supply the pre-training and post-training tasks for the second model configuration.

\paragraph{Side notes.}
\begin{itemize}
    \item Six tasks in the ARC AGI II evaluation split also appear in ARC AGI I. When adapting models pre-trained on ARC AGI I, the \texttt{arc-agi\_evaluation2clean\_challenges.json} split filters these duplicates to avoid contamination.
    \item The placeholder \texttt{arc-agi\_test\_challenges.json} split contains fewer test examples than the evaluation set. Inference on this set is roughly 33\% faster than the final competition rerun and can under-estimate runtime, risking notebook timeouts during submission tests.
\end{itemize}

All data is available in the Trelis TRM fork on Github \cite{TrelisResearchTRM}
\end{document}